\documentclass[nopdfa]{ceurart}

\usepackage{amsmath}

\usepackage{enumitem}
\usepackage{tabularx}
\usepackage{tikz}
\usepackage{xcolor}
\usetikzlibrary{positioning, arrows.meta, calc, fit, backgrounds}
\definecolor{tikzblue}{RGB}{31,119,180}
\definecolor{tikzlightblue}{RGB}{174,214,241}
\definecolor{tikzorange}{RGB}{255,127,14}
\definecolor{tikzlightorange}{RGB}{255,218,185}
\definecolor{tikzgreen}{RGB}{44,160,44}
\definecolor{tikzlightgreen}{RGB}{198,232,200}
\usepackage{pgfplots}
\pgfplotsset{compat=1.17}

\providecommand{\Description}[1]{}

\begin{document}

%% ---- CEUR rights + venue (replaces the ACM \acmConference block) ----
\copyrightyear{2026}
\copyrightclause{Copyright for this paper by its authors.
  Use permitted under Creative Commons License Attribution 4.0
  International (CC BY 4.0).}

\conference{Workshop on Recommenders in Tourism (RecTour 2026), September 28, 2026,
  co-located with the 20th ACM Conference on Recommender Systems,
  Minneapolis, Minnesota, USA.}

%% ---- Title ----
\title{A Multi-Source Ensemble Approach to Candidate Generation for
  Alternative Vacation Rental Property Recommendations}

%% ---- Authors (single-blind: real names shown) ----
\author[1]{Syed Mohammed Arshad Zaidi}[%
  email=syzaidi@expediagroup.com,
]
\cormark[1]
\author[1]{Eric Rincon}[%
  email=erincon@expediagroup.com,
]
\author[1]{Shayan Hassantabar}[%
  email=shassantabar@expediagroup.com,
]
\address[1]{Expedia Group, Austin, Texas, USA}
\cortext[1]{Corresponding author.}

\begin{abstract}
Alternative property recommendations play a critical role in vacation rental marketplaces, helping users discover relevant options when viewing a specific listing. However, generating high-quality candidate alternatives presents unique challenges: heterogeneous inventory, geographic constraints, rapid availability changes, and long-tail property distributions. We present a comprehensive study of candidate generation (CG) approaches for vacation rental alternatives, comparing collaborative filtering, shallow embeddings, and graph neural network (GNN) methods.

Our experiments on a large-scale vacation rental platform (over 2M active properties) show that a hybrid architecture combining item-based collaborative filtering with GNN-based retrieval improves Recall@300 by 14.8\% over the strongest baseline, by leveraging the complementary strengths of the two sources: collaborative filtering excels at early recall for properties with rich interaction history, while GNNs discover diverse, non-obvious alternatives and handle cold-start scenarios more effectively. As a component result, GNN-based embeddings alone substantially outperform shallow Hotel2Vec embeddings (48--68\% relative recall improvement across K), motivating their inclusion in the ensemble.

Crucially, we examine how CG-stage gains carry through to the downstream ranking stage, and find that a stronger candidate pool yields higher downstream ranking quality -- though attributing this effect cleanly is complicated by the coupling between candidate generation and ranker training. This \textit{recall-conversion gap} is an important consideration for practitioners deploying new retrieval methods in two-stage recommendation systems.
\end{abstract}

\begin{keywords}
  candidate generation \sep
  graph neural networks \sep
  recommendation systems \sep
  vacation rentals \sep
  two-stage retrieval \sep
  collaborative filtering \sep
  embeddings
\end{keywords}

\maketitle

%% =====================================================================
%% BODY -- carried over verbatim from the ACM version.
%% =====================================================================

\section{Introduction}
\begin{figure}[t]
\centering
\resizebox{0.72\columnwidth}{!}{%
\begin{tikzpicture}[
  font=\sffamily\footnotesize,
]

% ========== COLOR PALETTE ==========
\definecolor{brandNavy}{RGB}{27, 54, 93}
\definecolor{brandBlue}{RGB}{45, 95, 155}
\definecolor{brandTeal}{RGB}{64, 145, 158}
\definecolor{brandLight}{RGB}{142, 202, 206}

\definecolor{annotRef}{RGB}{180, 80, 60}
\definecolor{annotAlt}{RGB}{50, 135, 90}

\definecolor{cardBg}{RGB}{255,255,255}
\definecolor{textDark}{RGB}{45, 55, 65}
\definecolor{textGray}{RGB}{110, 120, 130}
\definecolor{borderGray}{RGB}{220, 225, 230}
\definecolor{bgLight}{RGB}{248, 250, 252}

\definecolor{ratingGreen}{RGB}{0, 135, 90}
\definecolor{accentCoral}{RGB}{235, 120, 100}

\definecolor{skyTop}{RGB}{165, 210, 235}
\definecolor{skyBottom}{RGB}{200, 225, 245}
\definecolor{oceanDeep}{RGB}{70, 140, 170}
\definecolor{oceanLight}{RGB}{120, 180, 200}
\definecolor{sandWarm}{RGB}{235, 215, 185}
\definecolor{sandLight}{RGB}{245, 235, 220}
\definecolor{palmDark}{RGB}{60, 120, 80}
\definecolor{palmLight}{RGB}{90, 155, 95}
\definecolor{roofWarm}{RGB}{175, 100, 75}
\definecolor{roofLight}{RGB}{200, 130, 100}
\definecolor{wallCream}{RGB}{255, 252, 245}
\definecolor{sunGlow}{RGB}{255, 220, 120}
\definecolor{accentGold}{RGB}{245, 180, 60}

% Define left margins
\def\refTextLeft{-3.85}
\def\cardTextLeft{-0.95}

% Shift for rating boxes (move right to align with text)
\def\refRatingShift{0.10}
\def\cardRatingShift{0.10}

% ========== MAIN PROPERTY IMAGE ==========
\begin{scope}
  \clip[rounded corners=4pt] (-4,2.6) rectangle (4,0.6);
  \fill[skyTop] (-4,2.6) rectangle (4,1.8);
  \fill[skyBottom] (-4,1.8) rectangle (4,1.4);
  \fill[sunGlow, opacity=0.3] (3.2,2.2) circle (0.5);
  \fill[sunGlow] (3.2,2.2) circle (0.28);
  \fill[oceanDeep] (-4,1.4) rectangle (4,1.1);
  \fill[oceanLight] (-4,1.1) rectangle (4,0.95);
  \fill[sandWarm] (-4,0.95) rectangle (4,0.6);
  
  \fill[wallCream] (-1.4,2.1) rectangle (1.4,1.25);
  \fill[roofWarm] (-1.6,2.25) -- (0,2.6) -- (1.6,2.25) -- (1.4,2.1) -- (-1.4,2.1) -- cycle;
  \foreach \wx in {-0.85, -0.05, 0.75} {
    \fill[brandLight!60] (\wx,1.9) rectangle (\wx+0.35,1.55);
    \fill[skyBottom!80] (\wx+0.02,1.88) rectangle (\wx+0.33,1.57);
  }
  \fill[roofLight] (-0.12,1.47) rectangle (0.12,1.25);
  
  \fill[palmDark] (-2.9,1.3) -- (-3.15,2.1) -- (-2.65,2.1) -- cycle;
  \fill[palmLight] (-2.75,1.45) -- (-3.1,2.2) -- (-2.5,2.0) -- cycle;
  \fill[roofLight!70] (-2.82,0.9) rectangle (-2.72,1.55);
  
  \fill[palmDark] (2.9,1.3) -- (2.65,2.1) -- (3.15,2.1) -- cycle;
  \fill[palmLight] (2.75,1.45) -- (2.5,2.0) -- (3.1,2.2) -- cycle;
  \fill[roofLight!70] (2.72,0.9) rectangle (2.82,1.55);
  
  \fill[accentCoral] (-2.1,1.65) -- (-2.4,1.2) -- (-1.8,1.2) -- cycle;
  \draw[roofLight!80, line width=1pt] (-2.1,0.85) -- (-2.1,1.45);
\end{scope}

% Photo dots
\foreach \x in {-0.15,0.05,0.25} {
  \fill[white, opacity=0.6] (\x,0.75) circle (0.05);
}
\fill[brandTeal] (0.05,0.75) circle (0.05);

% Main property details
\node[anchor=west, font=\sffamily\bfseries\small, text=textDark] at (\refTextLeft,0.28) {Oceanview Villa -- Beachfront Paradise};
\node[anchor=west, font=\scriptsize, text=textGray] at (\refTextLeft,0.0) {Miami Beach, Florida · 4 BR · Sleeps 8};

% Guest rating - SHIFTED RIGHT
\fill[ratingGreen, rounded corners=2pt] (\refTextLeft+\refRatingShift,-0.48) rectangle (\refTextLeft+\refRatingShift+0.42,-0.28);
\node[font=\tiny\bfseries, text=white] at (\refTextLeft+\refRatingShift+0.21,-0.38) {9.4};
\node[anchor=west, font=\scriptsize, text=textGray] at (\refTextLeft+\refRatingShift+0.5,-0.38) {127 reviews};

% Price and button - RIGHT SIDE
% Price - right side, on the address line (well above the button)
\node[anchor=east, font=\bfseries\small, text=textDark] at (3.9,-0.15) {\$425/night};

% Book Now button - right side, lower row
\fill[brandNavy, rounded corners=3pt] (2.35,-0.30) rectangle (3.75,-0.58);
\node[text=white, font=\tiny\bfseries] at (3.05,-0.44) {Book Now};

% ========== FLOW ARROW - longer shaft with smaller arrowhead ==========
\draw[textGray!80, line width=2pt, -{Triangle[length=5pt, width=4pt]}, line cap=round] (0,-0.67) -- (0,-0.97);

% Carousel header
\node[anchor=west, font=\sffamily\bfseries\small, text=textDark] at (\refTextLeft,-1.17) {Similar properties you may like};
\node[anchor=east, font=\scriptsize, text=brandBlue] at (3.9,-1.17) {View all $\rightarrow$};

% ===== Card 1: Beach House =====
\begin{scope}[shift={(-2.6,-0.05)}]
  \fill[black, opacity=0.04, rounded corners=3pt] (-1.1,-1.28) rectangle (1.2,-4.05);
  \fill[cardBg, rounded corners=3pt, draw=borderGray, line width=0.5pt] (-1.15,-1.33) rectangle (1.15,-4.08);
  
  \begin{scope}
    \clip[rounded corners=2pt] (-1.05,-1.43) rectangle (1.05,-2.55);
    \fill[skyTop] (-1.05,-1.43) rectangle (1.05,-1.78);
    \fill[oceanDeep] (-1.05,-1.78) rectangle (1.05,-2.1);
    \fill[sandWarm] (-1.05,-2.1) rectangle (1.05,-2.55);
    \fill[wallCream] (-0.35,-1.62) rectangle (0.35,-2.05);
    \fill[roofWarm] (-0.45,-1.51) -- (0,-1.36) -- (0.45,-1.51) -- (0.35,-1.62) -- (-0.35,-1.62) -- cycle;
    \fill[skyBottom!70] (-0.18,-1.72) rectangle (0.0,-1.89);
    \fill[skyBottom!70] (0.08,-1.72) rectangle (0.26,-1.89);
    \fill[palmDark] (0.65,-1.66) -- (0.48,-1.43) -- (0.82,-1.43) -- cycle;
  \end{scope}
  
  \node[anchor=west, font=\scriptsize\bfseries, text=textDark] at (\cardTextLeft,-2.75) {Beach House};
  \node[anchor=west, font=\tiny, text=textGray] at (\cardTextLeft,-2.95) {Miami Beach};
  
  \fill[ratingGreen, rounded corners=1pt] (\cardTextLeft+\cardRatingShift,-3.30) rectangle (\cardTextLeft+\cardRatingShift+0.30,-3.14);
  \node[font=\tiny\bfseries, text=white] at (\cardTextLeft+\cardRatingShift+0.15,-3.22) {9.2};
  \node[anchor=west, font=\tiny, text=textGray] at (\cardTextLeft+\cardRatingShift+0.36,-3.22) {45 reviews};
  
  \node[anchor=west, font=\scriptsize\bfseries, text=textDark] at (\cardTextLeft,-3.52) {\$380/night};
  
  \fill[brandTeal!15, rounded corners=2pt] (-0.95,-3.72) rectangle (0.95,-3.92);
  \node[text=brandTeal, font=\tiny\bfseries] at (0,-3.82) {View Property};
\end{scope}

% ===== Card 2: Coastal Retreat =====
\begin{scope}[shift={(0,-0.05)}]
  \fill[black, opacity=0.04, rounded corners=3pt] (-1.1,-1.28) rectangle (1.2,-4.05);
  \fill[cardBg, rounded corners=3pt, draw=borderGray, line width=0.5pt] (-1.15,-1.33) rectangle (1.15,-4.08);
  
  \begin{scope}
    \clip[rounded corners=2pt] (-1.05,-1.43) rectangle (1.05,-2.55);
    \fill[skyBottom] (-1.05,-1.43) rectangle (1.05,-1.65);
    \fill[bgLight] (-0.55,-1.53) rectangle (0.55,-2.28);
    \fill[brandLight!50] (-0.4,-1.62) rectangle (-0.12,-1.8);
    \fill[brandLight!50] (0.12,-1.62) rectangle (0.4,-1.8);
    \fill[brandLight!50] (-0.4,-1.9) rectangle (-0.12,-2.08);
    \fill[brandLight!50] (0.12,-1.9) rectangle (0.4,-2.08);
    \fill[brandTeal!45] (-0.75,-2.3) rectangle (0.75,-2.5);
    \fill[sandLight] (-0.78,-2.22) rectangle (0.78,-2.3);
  \end{scope}
  
  \node[anchor=west, font=\scriptsize\bfseries, text=textDark] at (\cardTextLeft,-2.75) {Coastal Retreat};
  \node[anchor=west, font=\tiny, text=textGray] at (\cardTextLeft,-2.95) {Sunny Isles};
  
  \fill[ratingGreen, rounded corners=1pt] (\cardTextLeft+\cardRatingShift,-3.30) rectangle (\cardTextLeft+\cardRatingShift+0.30,-3.14);
  \node[font=\tiny\bfseries, text=white] at (\cardTextLeft+\cardRatingShift+0.15,-3.22) {8.8};
  \node[anchor=west, font=\tiny, text=textGray] at (\cardTextLeft+\cardRatingShift+0.36,-3.22) {23 reviews};
  
  \node[anchor=west, font=\scriptsize\bfseries, text=textDark] at (\cardTextLeft,-3.52) {\$295/night};
  
  \fill[brandTeal!15, rounded corners=2pt] (-0.95,-3.72) rectangle (0.95,-3.92);
  \node[text=brandTeal, font=\tiny\bfseries] at (0,-3.82) {View Property};
\end{scope}

% ===== Card 3: Bay Villa =====
\begin{scope}[shift={(2.6,-0.05)}]
  \fill[black, opacity=0.04, rounded corners=3pt] (-1.1,-1.28) rectangle (1.2,-4.05);
  \fill[cardBg, rounded corners=3pt, draw=borderGray, line width=0.5pt] (-1.15,-1.33) rectangle (1.15,-4.08);
  
  \begin{scope}
    \clip[rounded corners=2pt] (-1.05,-1.43) rectangle (1.05,-2.55);
    \fill[skyTop] (-1.05,-1.43) rectangle (1.05,-1.62);
    \fill[accentGold!25] (-1.05,-1.43) rectangle (1.05,-1.54);
    \fill[sunGlow!60] (0.58,-1.5) circle (0.1);
    \fill[oceanLight] (-1.05,-1.62) rectangle (1.05,-1.95);
    \fill[palmLight!35] (-1.05,-1.95) rectangle (1.05,-2.55);
    \fill[wallCream] (-0.42,-1.78) rectangle (0.42,-2.22);
    \fill[roofWarm] (-0.52,-1.67) -- (0,-1.52) -- (0.52,-1.67) -- (0.42,-1.78) -- (-0.42,-1.78) -- cycle;
    \fill[skyBottom!60] (-0.28,-1.9) rectangle (-0.08,-2.05);
    \fill[skyBottom!60] (0.08,-1.9) rectangle (0.28,-2.05);
    \fill[sandLight] (-0.48,-2.22) rectangle (0.48,-2.35);
  \end{scope}
  
  \node[anchor=west, font=\scriptsize\bfseries, text=textDark] at (\cardTextLeft,-2.75) {Bay Villa};
  \node[anchor=west, font=\tiny, text=textGray] at (\cardTextLeft,-2.95) {Key Biscayne};
  
  \fill[ratingGreen, rounded corners=1pt] (\cardTextLeft+\cardRatingShift,-3.30) rectangle (\cardTextLeft+\cardRatingShift+0.30,-3.14);
  \node[font=\tiny\bfseries, text=white] at (\cardTextLeft+\cardRatingShift+0.15,-3.22) {9.6};
  \node[anchor=west, font=\tiny, text=textGray] at (\cardTextLeft+\cardRatingShift+0.36,-3.22) {89 reviews};
  
  \node[anchor=west, font=\scriptsize\bfseries, text=textDark] at (\cardTextLeft,-3.52) {\$445/night};
  
  \fill[brandTeal!15, rounded corners=2pt] (-0.95,-3.72) rectangle (0.95,-3.92);
  \node[text=brandTeal, font=\tiny\bfseries] at (0,-3.82) {View Property};
\end{scope}

% ========== CAROUSEL ARROWS ==========
\fill[white, draw=borderGray, line width=0.5pt, rounded corners=2pt] (-3.95,-2.55) rectangle (-3.55,-2.95);
\draw[textGray, line width=1.2pt, line cap=round, line join=round] (-3.7,-2.65) -- (-3.82,-2.75) -- (-3.7,-2.85);

\fill[white, draw=borderGray, line width=0.5pt, rounded corners=2pt] (3.55,-2.55) rectangle (3.95,-2.95);
\draw[textGray, line width=1.2pt, line cap=round, line join=round] (3.7,-2.65) -- (3.82,-2.75) -- (3.7,-2.85);

% ========== ANNOTATIONS ==========
\draw[annotRef, thick, dashed, rounded corners=3pt] (-4.08,2.75) rectangle (4.08,-0.72);
\node[fill=annotRef, text=white, font=\tiny\bfseries, rounded corners=2pt, inner sep=3pt] 
  at (4.08,2.75) [anchor=south east] {Reference Property $P_{ref}$};

\draw[annotAlt, thick, dashed, rounded corners=3pt] (-4.08,-1.02) rectangle (4.08,-4.23);
\node[fill=annotAlt, text=white, font=\tiny\bfseries, rounded corners=2pt, inner sep=3pt]
  at (0,-4.23) [anchor=north] {Alternative Recommendations $\{P_1, P_2, ..., P_K\}$};

\end{tikzpicture}%
}%
\Description{A mockup of a vacation rental platform interface showing a reference property listing at the top with an illustrated beachfront villa, guest rating badge, and price per night. Below is a carousel of three similar alternative property recommendations, each with property image, location, guest rating, and price.}
\caption{Alternative property recommendations on a vacation rental platform. When a user views a reference property $P_{ref}$, the system generates a carousel of $K$ alternative properties $\{P_1, P_2, ..., P_K\}$ that match the user's destination and preferences.}
\label{fig:ui_mockup}
\end{figure}
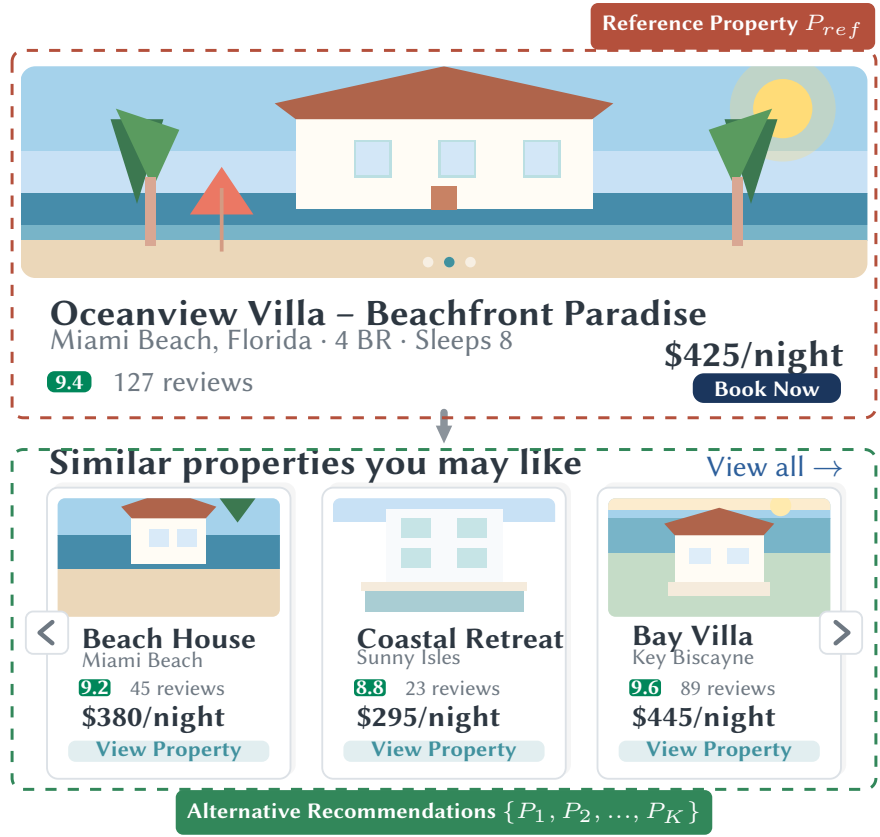

Recommendation in travel and tourism differs from general item recommendation in ways that directly shape system design: user intent is anchored to a destination and specific travel dates, demand is strongly seasonal, and the bookable inventory shifts continually as properties are listed, reserved, and become unavailable. Vacation rental marketplaces are a particularly demanding instance of this setting. When travelers browse such platforms, they typically view several properties before booking, so presenting relevant \emph{alternative} properties--similar listings that may better match a traveler's needs--is central to both user experience and conversion (Fig.~\ref{fig:ui_mockup}). Unlike traditional hotel recommendation, where inventory is relatively standardized, vacation rental marketplaces face compounding challenges: inventory is highly heterogeneous and largely unique per listing--from studio apartments to multi-bedroom villas--subject to strict geographic constraints tied to destination intent and to rapidly changing availability, and heavily long-tailed. A continual stream of newly onboarded properties, moreover, makes cold-start a standing condition rather than an occasional edge case, which is a central concern for retrieval in this domain.

Candidate generation (CG), the first stage of modern two-stage recommendation systems~\cite{covington2016deep}, must address these challenges while retrieving a diverse set of relevant alternatives from catalogs containing millions of properties. Existing approaches each have significant limitations when applied individually. Collaborative filtering methods like item-based k-nearest neighbors (IBKNN) excel for properties with rich interaction history but struggle with cold-start listings. Embedding methods such as Hotel2Vec~\cite{sadeghian2019hotel2vec} capture co-occurrence patterns efficiently but learn only symmetric, single-hop relationships that miss complex patterns in user behavior. Distance-based retrieval ensures geographic relevance but lacks semantic understanding of property similarity.

Graph neural networks (GNNs) offer a principled way to address these limitations. By modeling properties, users, and interactions as a heterogeneous graph, GNNs can learn asymmetric relationships, capture multi-hop patterns via message passing, and incorporate rich property features during representation learning. This motivates our investigation of GNN-based candidate generation for vacation rental recommendations.

Concretely, we frame candidate generation as link prediction on a heterogeneous property-user graph, with reference and candidate properties as distinct node types so that the model can learn directional ``A is an alternative to B'' relationships. The graph integrates user-property interactions (a behavioral signal) and validated booking-derived alternative pairs (a supervised signal), and property attributes spanning structural, location, policy, quality, and amenity features are aggregated through message passing so that even properties with sparse interaction history can be embedded into a sensible region of the candidate space. 
We use this GNN as one retrieval source in a multi-source candidate
generator alongside IBKNN, integrating its learned representations with
collaborative-filtering signals to exploit their complementary strengths:
IBKNN dominates early recall on properties with rich behavioral signal,
while the GNN expands late-recall coverage and represents cold-start
listings more effectively. This design directly targets the two failure
modes of single-source CG identified above \textendash{} popularity bias
for collaborative filtering and shallow, symmetric similarity for embedding
methods \textendash{} and we use the resulting candidate pool to study how
candidate-generation gains carry through to the downstream ranking stage.

In this work, we present a comprehensive study of candidate generation approaches for alternative vacation rental recommendations. We make the following contributions:

\begin{enumerate}
    \item \textbf{Hybrid architecture with complementary strengths.} We propose combining IBKNN's early-recall strength with GNN's ability to discover diverse, non-obvious alternatives, achieving +14.8\% improvement at Recall@300 compared to the strongest baseline on a large-scale vacation rental platform with over 2M properties.
    
    \item \textbf{Systematic evaluation of CG methods.} We compare collaborative filtering (IBKNN), shallow embeddings (Hotel2Vec), and graph neural networks (GNN), demonstrating that GNN-based embeddings achieve 48--68\% relative improvement in recall over shallow embedding baselines and motivating their inclusion in the hybrid.
    
    \item \textbf{Connecting CG gains to downstream ranking.} We measure how candidate-pool quality affects downstream ranking, finding that the hybrid pool yields higher Booking NDCG@5 than the baseline pool, and we discuss the recall-conversion gap that complicates clean attribution of CG improvements to end-to-end gains.
\end{enumerate}

Our findings provide actionable insights for practitioners: a stronger candidate pool can improve downstream ranking, but isolating the CG contribution from confounding ranker-training effects requires careful experimental design.

The remainder of this paper is organized as follows. Section~\ref{sec:related_work} reviews related work on candidate generation and GNN-based recommendations.\ Section~\ref{sec:background} provides background on the embedding methods evaluated. Section~\ref{sec:methodology} describes our multi-source CG architecture. Section~\ref{sec:experiments} details our experimental setup, and Section~\ref{sec:results} presents results. We discuss implications and limitations in Section~\ref{sec:discussion} and conclude in Section~\ref{sec:conclusion}.

%================================================================
\section{Related Work}\label{sec:related_work}
\textbf{Two-Stage Recommendation Systems.} Modern recommendation systems employ a two-stage architecture consisting of candidate generation (retrieval) followed by ranking~\cite{covington2016deep}. This paradigm addresses the computational challenge of scoring millions of items by first narrowing the candidate pool to $O(1000)$ candidates, which are then ranked to surface $O(10)$ final recommendations. Two-tower architectures, where separate encoders learn query and item representations for efficient approximate nearest neighbor search, have become the industry standard~\cite{yi2019sampling, osowska2025suggest}. Huang et al.~\cite{huang2020embedding} proposed methods for learning embeddings that transfer across retrieval and ranking stages, addressing the distribution mismatch between stages - a challenge directly relevant to our work.

\textbf{Multi-Source Candidate Generation.} Industrial systems increasingly combine multiple retrieval sources for robustness and coverage. LinkedIn's People You May Know system, for example, combines graph-based sources (e.g., triangle closing on the connection graph), embedding-based retrieval, and heuristic sources to process hundreds of billions of potential connections daily. Recent industrial work on item-to-item retrieval similarly explores multi-task and multi-head architectures to jointly optimize recall and relevance at billion-user scale~\cite{zhang2025optimizing}. Industrial deployments also report that careful candidate-generation design \textendash{} including bias-aware sampling \textendash{} is critical for downstream quality~\cite{virani2020lessons}. The rationale for multi-source architectures is threefold: (1) no single algorithm captures all aspects of user behavior; (2) multiple sources ensure broader coverage across inventory; and (3) system performance degrades gracefully if one source fails. The challenge of translating retrieval improvements into end-to-end gains - what we term the \textit{recall-conversion gap} - has received limited attention in the literature, though practitioners frequently encounter this issue when deploying new retrieval methods.

\textbf{Graph-Based Candidate Generation.} Graph methods model user-item relationships as network structures for candidate retrieval. Pinterest's Pixie system~\cite{eksombatchai2018pixie} performs biased random walks on a bipartite pin-board graph, with innovations including multi-query weighting, multi-hit boosting for pins reachable from multiple query nodes, and early stopping for efficiency.\ PinSage~\cite{ying2018graph} extended this to GNN-based embeddings, demonstrating scalability to billions of nodes. More recently, OmniSage~\cite{badrinath2025omnisage} unified graph neural networks with content-based models and user sequence models through multiple contrastive learning tasks on heterogeneous graphs with billions of nodes, achieving a 2.5\% increase in sitewide engagement at Pinterest. Chen et al.~\cite{chen2024macro} address the scalability challenges of GNNs for online billion-scale recommender systems through macro-level graph constructions, demonstrating practical deployment at industrial scale. LightGCN~\cite{he2020lightgcn} simplified graph convolutions for collaborative filtering, showing that neighborhood aggregation is the key component. Wu et al.~\cite{wu2022graph} surveyed GNN-based recommendation methods, highlighting advantages for multi-hop relationships and cold-start handling through feature propagation.

\textbf{Embedding-Based Retrieval with Side Information.} Alibaba's billion-scale recommendation system~\cite{wang2018billion} learns item embeddings from user behavior graphs using random walk and skip-gram methods. Their Graph Embedding with Side Information (GES) addresses cold-start by incorporating category, brand, and shop features, while Enhanced GES (EGES) uses attention-weighted aggregation to learn the importance of different side information types. This demonstrates that combining behavioral co-occurrence signals with explicit item features significantly improves embedding quality - an insight that motivates our hybrid architecture combining collaborative filtering with feature-rich GNN embeddings.

\textbf{Lodging and Travel Recommendations.} Travel recommendation presents unique challenges including geographic constraints, temporal availability, and inventory heterogeneity. Hotel2Vec~\cite{sadeghian2019hotel2vec} applied Word2Vec-style embedding techniques to learn hotel representations from booking sequences. Our work differs by focusing on the alternative property use case and systematically studying the gap between retrieval recall and end-to-end recommendation quality -- providing insights applicable beyond the travel domain to any two-stage system with heterogeneous inventory.

%================================================================
\section{Background}\label{sec:background}
This section provides background on the candidate generation approaches evaluated in this work.

\subsection{Hotel2Vec: Attribute-Enriched Property Embeddings}
Hotel2Vec~\cite{sadeghian2019hotel2vec} learns dense property representations by combining user click sessions with structured property attributes. The model extends the skip-gram paradigm: it treats user sessions as sequences and predicts context properties given a target property. Crucially, Hotel2Vec enriches click embeddings with additional inputs including geographic coordinates, star ratings, user ratings, and amenity features. These sub-embeddings are concatenated and fused through a neural layer to produce a unified representation.

The resulting embeddings capture co-view patterns---properties frequently viewed together are embedded nearby---while the attribute fusion helps address cold-start cases for new properties. For candidate generation, we retrieve the embedding for a reference property $P_{ref}$ and return the top-$K$ nearest neighbors from a FAISS index using cosine similarity.

Despite incorporating attributes, Hotel2Vec has notable limitations for our task: (1) single-hop context that cannot capture multi-hop relationships (e.g., properties connected through shared audiences); (2) symmetric similarity that treats ``A is alternative to B'' identically to ``B is alternative to A''; and (3) no iterative neighborhood aggregation to propagate information across the property graph.

\subsection{GNN-Based Property Embeddings}
Our GNN-based approach addresses these limitations by modeling the recommendation task as link prediction on a heterogeneous graph. The graph contains multiple node types (reference properties, candidate properties, users, interactions, and ground truth labels) connected by typed edges representing user-property interactions and validated alternative relationships.

A key design choice is separating properties into left-hand side (LHS) and right-hand side (RHS) nodes, enabling the model to learn \textit{asymmetric relationships} -- recognizing that ``property A is a good alternative to B'' does not necessarily imply the reverse. The GNN is trained to predict which RHS properties will serve as validated alternatives for each LHS property.

The GNN learns representations through iterative message passing, where nodes aggregate information from their neighbors across multiple hops. This captures both direct relationships and transitive patterns (e.g., properties connected through shared user audiences). Unlike Hotel2Vec, the GNN explicitly incorporates property features during message passing, including structural attributes (bedrooms, capacity), location, policies, quality signals (ratings, reviews), and 62 amenity indicators.

After training, the model produces two distinct embeddings per property: an LHS embedding (query representation) and an RHS embedding (target representation). Candidate generation uses asymmetric dot-product similarity: $\text{sim}(P_{ref}, P_{cand}) = e_{LHS}(P_{ref}) \cdot e_{RHS}(P_{cand})$.

\begin{figure}[t]
\centering
\resizebox{0.95\columnwidth}{!}{%
\begin{tikzpicture}[
    node distance=1cm and 2cm,
    header/.style={
        rectangle,
        rounded corners=2pt,
        minimum width=3.2cm,
        minimum height=0.45cm,
        fill=#1!30,
        font=\small\bfseries,
        inner sep=2pt
    },
    body/.style={
        rectangle,
        draw=#1!60,
        rounded corners=3pt,
        minimum width=3.4cm,
        fill=#1!8,
        font=\small,
        inner sep=6pt,
        align=left
    },
    conn/.style={draw=gray!50, thick},
    asymm/.style={draw=red!70, thick, dashed, <->}
]

% Users entity (gray)
\node[body=gray] (users) {
    \phantom{\textbf{users}}\\[2pt]
    PK: device\_user\_agent\_id
};
\node[header=gray, anchor=north] at ([yshift=-2pt]users.north) {users};

% Interactions entity (purple)
\node[body=violet, right=2cm of users, yshift=0.4cm] (interactions) {
    \phantom{\textbf{interactions}}\\[2pt]
    T: event\_date\\
    FK: device\_user\_agent\_id\\
    FK: product\_id
};
\node[header=violet, anchor=north] at ([yshift=-2pt]interactions.north) {interactions};

% hotels_lhs entity (blue)
\node[body=blue, below=1cm of users] (lhs) {
    \phantom{\textbf{hotels\_lhs}}\\[2pt]
    PK: eg\_property\_id
};
\node[header=blue, anchor=north] at ([yshift=-2pt]lhs.north) {hotels\_lhs};

% hotels_rhs entity (teal)
\node[body=teal, below=1cm of lhs] (rhs) {
    \phantom{\textbf{hotels\_rhs}}\\[2pt]
    PK: eg\_property\_id
};
\node[header=teal, anchor=north] at ([yshift=-2pt]rhs.north) {hotels\_rhs};

% Labels entity (orange)
\node[body=orange, right=2cm of rhs, yshift=0.4cm] (labels) {
    \phantom{\textbf{labels}}\\[2pt]
    T: event\_date\\
    FK: reference\_id\\
    FK: ground\_truth\_id
};
\node[header=orange, anchor=north] at ([yshift=-2pt]labels.north) {labels};

% Connections with curved lines
\draw[conn] (users.east) to[out=0, in=180] (interactions.west);
\draw[conn] (lhs.east) to[out=0, in=210] (interactions.south west);
\draw[conn] (rhs.east) to[out=0, in=180] (labels.west);
\draw[conn] (lhs.east) to[out=0, in=135] (labels.north west);

% Asymmetric relationship between lhs and rhs
\draw[asymm] (lhs.south) -- (rhs.north) 
    node[midway, right, font=\scriptsize, text=red!70, xshift=2pt] {asymmetric};

\end{tikzpicture}%
}
\Description{A diagram showing five connected entity boxes representing the heterogeneous graph schema: users, interactions, hotels_lhs, hotels_rhs, and labels. Arrows show foreign key relationships between entities, and a dashed red line indicates asymmetric similarity learning between LHS and RHS property nodes.}
\caption{Heterogeneous graph schema for GNN training. Properties are split into LHS (reference) and RHS (candidate) nodes to learn asymmetric relationships.}
\label{fig:graph_schema}
\end{figure}
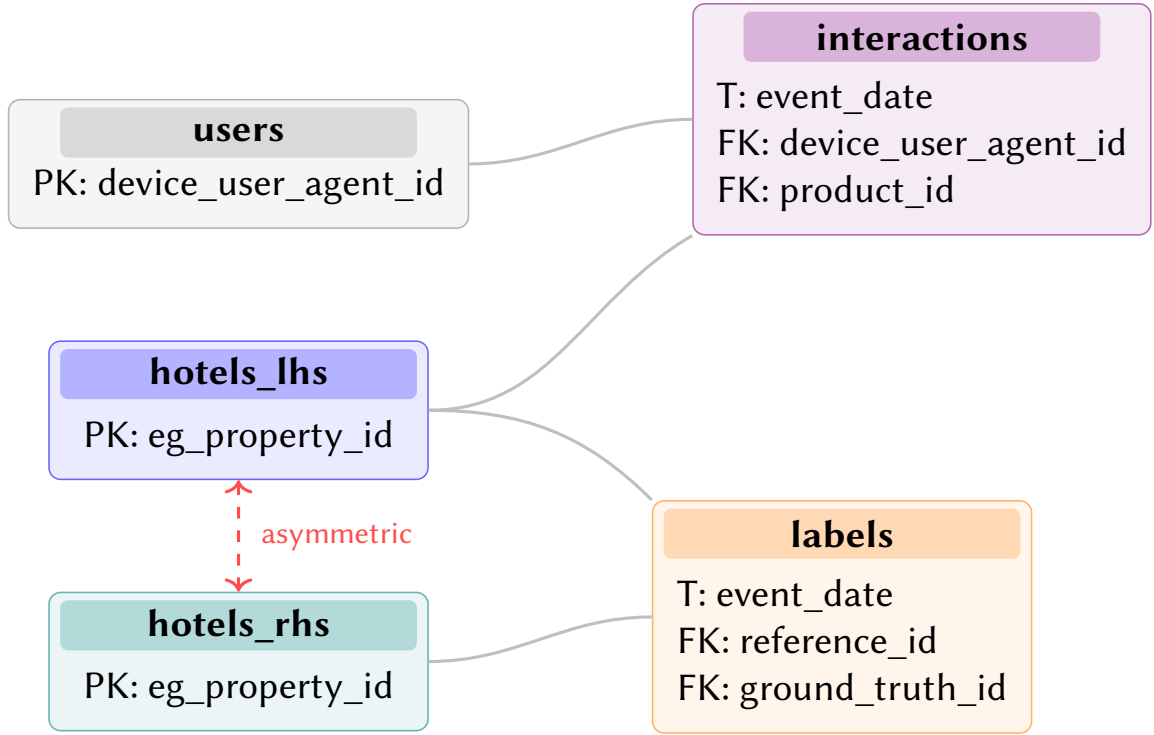

Figure~\ref{fig:graph_schema} illustrates the heterogeneous graph schema. The \texttt{users} node captures device-level user identifiers, while \texttt{interactions} records user-property engagements with timestamps. Properties are represented twice: \texttt{hotels\_lhs} serves as the query (reference) representation and \texttt{hotels\_rhs} as the target (candidate) representation, enabling the model to learn asymmetric similarity. The \texttt{labels} node contains ground truth alternative pairs derived from booking conversions, providing supervised signal for link prediction training.

\subsection{Baseline System}
The baseline candidate generation system combines multiple sources: (1) IBKNN -- item-based collaborative filtering using Jaccard similarity on audience overlap; (2) Hotel2Vec -- embedding similarity as described above; and (3) geographic filtering for destination relevance. Candidates are merged via sequential deduplication with source priority, then scored by a downstream ranking model trained on historical user engagement.

\subsection{Summary: Complementary Strengths}
Table~\ref{tab:cg_comparison} summarizes the strengths and limitations of each approach.

\begin{table}[!htbp]
\caption{Comparison of candidate generation approaches.}
\label{tab:cg_comparison}
\small
\begin{tabularx}{\columnwidth}{lXX}
\toprule
\textbf{Approach} & \textbf{Strengths} & \textbf{Limitations} \\
\midrule
IBKNN & Strong early recall; interpretable; fast & Requires interaction history; popularity bias \\
\addlinespace
Hotel2Vec & Efficient FAISS retrieval; captures co-occurrence & Shallow; symmetric; cold-start issues \\
\addlinespace
GNN & Asymmetric; explicit features; multi-hop; better cold-start & Higher computational cost \\
\bottomrule
\end{tabularx}
\end{table}

%================================================================
\section{Methodology}\label{sec:methodology}
This section describes our multi-source candidate generation architecture and the strategies for combining candidates from heterogeneous sources.

\subsection{Task Definition}
We formalize the alternative property recommendation task as follows. Given a reference property $P_{ref}$ that a user is currently viewing, the goal is to generate a candidate set $C = \{(P_1, s_1), ..., (P_K, s_K)\}$ of $K$ alternative properties with relevance scores. Candidates must satisfy three constraints: (1) \textit{geographic relevance}--properties should be within the user's intended destination; (2) \textit{feature compatibility}--candidates should match core requirements (e.g., pet-friendly, minimum bedrooms); and (3) \textit{availability}--properties must be active and bookable.

\subsection{Multi-Source Architecture}
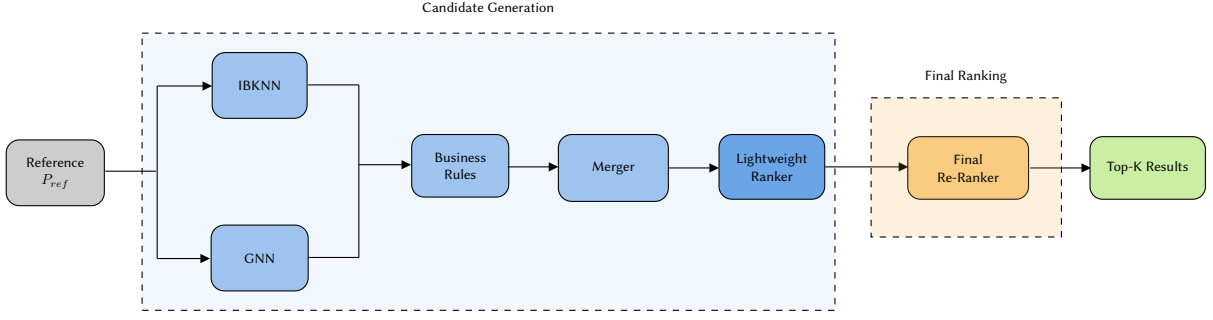
\begin{figure*}[t]
\centering
\resizebox{\linewidth}{!}{%
\tikzset{every picture/.style={line width=0.75pt}} %set default line width to 0.75pt        

\begin{tikzpicture}[x=0.75pt,y=0.75pt,yscale=-1,xscale=1]

%Shape: Rectangle - Candidate Generation box
\draw  [fill={rgb, 255:red, 74; green, 144; blue, 226 }  ,fill opacity=0.07 ][dash pattern={on 4.5pt off 4.5pt}] (147,162) -- (789.5,162) -- (789.5,419) -- (147,419) -- cycle ;

%Rounded Rect - Reference
\draw  [fill={rgb, 255:red, 155; green, 155; blue, 155 }  ,fill opacity=0.5 ] (21,273.2) .. controls (21,266.46) and (26.46,261) .. (33.2,261) -- (99.8,261) .. controls (106.54,261) and (112,266.46) .. (112,273.2) -- (112,309.8) .. controls (112,316.54) and (106.54,322) .. (99.8,322) -- (33.2,322) .. controls (26.46,322) and (21,316.54) .. (21,309.8) -- cycle ;

%Straight Lines - from Reference
\draw    (113,290) -- (160,290) ;
\draw    (161,211) -- (161,371) ;
\draw    (161,211) -- (209,211) ;
\draw [shift={(212,211)}, rotate = 180] [fill={rgb, 255:red, 0; green, 0; blue, 0 }  ][line width=0.08]  [draw opacity=0] (8.93,-4.29) -- (0,0) -- (8.93,4.29) -- cycle    ;
\draw    (161,371) -- (207,371) ;
\draw [shift={(210,371)}, rotate = 180] [fill={rgb, 255:red, 0; green, 0; blue, 0 }  ][line width=0.08]  [draw opacity=0] (8.93,-4.29) -- (0,0) -- (8.93,4.29) -- cycle    ;

%Rounded Rect - Kumo GNN
\draw  [fill={rgb, 255:red, 74; green, 144; blue, 226 }  ,fill opacity=0.5 ] (211,352.2) .. controls (211,345.46) and (216.46,340) .. (223.2,340) -- (288.8,340) .. controls (295.54,340) and (301,345.46) .. (301,352.2) -- (301,388.8) .. controls (301,395.54) and (295.54,401) .. (288.8,401) -- (223.2,401) .. controls (216.46,401) and (211,395.54) .. (211,388.8) -- cycle ;

%Straight Lines - from IBKNN and Kumo GNN to junction
\draw    (301,210) -- (348,210) ;
\draw    (301,370) -- (348,370) ;
\draw    (348,210) -- (348,370) ;
\draw    (348,284) -- (393.5,284) ;
\draw [shift={(396.5,284)}, rotate = 180] [fill={rgb, 255:red, 0; green, 0; blue, 0 }  ][line width=0.08]  [draw opacity=0] (8.93,-4.29) -- (0,0) -- (8.93,4.29) -- cycle    ;

%Rounded Rect - Business Rules
\draw  [fill={rgb, 255:red, 74; green, 144; blue, 226 }  ,fill opacity=0.5 ] (397,267.6) .. controls (397,261.19) and (402.19,256) .. (408.6,256) -- (475.4,256) .. controls (481.81,256) and (487,261.19) .. (487,267.6) -- (487,302.4) .. controls (487,308.81) and (481.81,314) .. (475.4,314) -- (408.6,314) .. controls (402.19,314) and (397,308.81) .. (397,302.4) -- cycle ;

%Straight Lines - to Merger
\draw    (486,286) -- (530,286) ;
\draw [shift={(533,286)}, rotate = 180] [fill={rgb, 255:red, 0; green, 0; blue, 0 }  ][line width=0.08]  [draw opacity=0] (8.93,-4.29) -- (0,0) -- (8.93,4.29) -- cycle    ;

%Rounded Rect - Merger
\draw  [fill={rgb, 255:red, 74; green, 144; blue, 226 }  ,fill opacity=0.5 ] (533,268.4) .. controls (533,261.55) and (538.55,256) .. (545.4,256) -- (622.6,256) .. controls (629.45,256) and (635,261.55) .. (635,268.4) -- (635,305.6) .. controls (635,312.45) and (629.45,318) .. (622.6,318) -- (545.4,318) .. controls (538.55,318) and (533,312.45) .. (533,305.6) -- cycle ;

%Straight Lines - to Lightweight Ranker
\draw    (635,287) -- (679,287) ;
\draw [shift={(682,287)}, rotate = 180] [fill={rgb, 255:red, 0; green, 0; blue, 0 }  ][line width=0.08]  [draw opacity=0] (8.93,-4.29) -- (0,0) -- (8.93,4.29) -- cycle    ;

%Rounded Rect - Lightweight Ranker
\draw  [fill={rgb, 255:red, 74; green, 144; blue, 226 }  ,fill opacity=0.8 ] (682,268) .. controls (682,261.37) and (687.37,256) .. (694,256) -- (768,256) .. controls (774.63,256) and (780,261.37) .. (780,268) -- (780,304) .. controls (780,310.63) and (774.63,316) .. (768,316) -- (694,316) .. controls (687.37,316) and (682,310.63) .. (682,304) -- cycle ;

%Rounded Rect - pAction Ranker
\draw  [fill={rgb, 255:red, 245; green, 166; blue, 35 }  ,fill opacity=0.5 ] (857,269.6) .. controls (857,263.19) and (862.19,258) .. (868.6,258) -- (957.4,258) .. controls (963.81,258) and (969,263.19) .. (969,269.6) -- (969,304.4) .. controls (969,310.81) and (963.81,316) .. (957.4,316) -- (868.6,316) .. controls (862.19,316) and (857,310.81) .. (857,304.4) -- cycle ;

%Straight Lines - to Top-K Results
\draw    (970.5,287) -- (1024,287) ;
\draw [shift={(1027,287)}, rotate = 180] [fill={rgb, 255:red, 0; green, 0; blue, 0 }  ][line width=0.08]  [draw opacity=0] (8.93,-4.29) -- (0,0) -- (8.93,4.29) -- cycle    ;

%Rounded Rect - Top-K Results
\draw  [fill={rgb, 255:red, 184; green, 233; blue, 134 }  ,fill opacity=0.75 ] (1026,269.6) .. controls (1026,263.19) and (1031.19,258) .. (1037.6,258) -- (1121.4,258) .. controls (1127.81,258) and (1133,263.19) .. (1133,269.6) -- (1133,304.4) .. controls (1133,310.81) and (1127.81,316) .. (1121.4,316) -- (1037.6,316) .. controls (1031.19,316) and (1026,310.81) .. (1026,304.4) -- cycle ;

%Straight Lines - Lightweight Ranker to pAction Ranker
\draw    (780.5,286) -- (854.5,286) ;
\draw [shift={(857.5,286)}, rotate = 180] [fill={rgb, 255:red, 0; green, 0; blue, 0 }  ][line width=0.08]  [draw opacity=0] (8.93,-4.29) -- (0,0) -- (8.93,4.29) -- cycle    ;

%Shape: Rectangle - Final Ranking box
\draw  [fill={rgb, 255:red, 245; green, 166; blue, 35 }  ,fill opacity=0.15 ][dash pattern={on 4.5pt off 4.5pt}] (823,222) -- (999.5,222) -- (999.5,351) -- (823,351) -- cycle ;

%Rounded Rect - IBKNN
\draw  [fill={rgb, 255:red, 74; green, 144; blue, 226 }  ,fill opacity=0.5 ] (212,192.2) .. controls (212,185.46) and (217.46,180) .. (224.2,180) -- (287.8,180) .. controls (294.54,180) and (300,185.46) .. (300,192.2) -- (300,228.8) .. controls (300,235.54) and (294.54,241) .. (287.8,241) -- (224.2,241) .. controls (217.46,241) and (212,235.54) .. (212,228.8) -- cycle ;

% %Dashed Lines - GNN features
% \draw  [dash pattern={on 4.5pt off 4.5pt}]  (253.5,401) -- (253.5,431) ;
% \draw  [dash pattern={on 4.5pt off 4.5pt}]  (253.5,431) -- (909.5,431) ;
% \draw  [dash pattern={on 4.5pt off 4.5pt}]  (908.53,319) -- (909.5,431) ;
% \draw [shift={(908.5,316)}, rotate = 89.5] [fill={rgb, 255:red, 0; green, 0; blue, 0 }  ][line width=0.08]  [draw opacity=0] (8.93,-4.29) -- (0,0) -- (8.93,4.29) -- cycle    ;

% Text Node - Reference (centered)
\draw (66.5,291.5) node [anchor=center][inner sep=0.75pt] [align=center] {{\small Reference}\\$P_{ref}$};

% Text Node - IBKNN (centered)
\draw (256,210.5) node [anchor=center][inner sep=0.75pt] [align=center] {IBKNN};

% Text Node - Kumo GNN (centered)
\draw (256,370.5) node [anchor=center][inner sep=0.75pt] [align=center] {GNN};

% Text Node - Business Rules (centered)
\draw (442,285) node [anchor=center][inner sep=0.75pt] [align=center] {Business\\Rules};

% Text Node - Merger (centered)
\draw (584,287) node [anchor=center][inner sep=0.75pt] [align=center] {Merger};

% Text Node - Lightweight Ranker (centered)
\draw (731,286) node [anchor=center][inner sep=0.75pt] [align=center] {Lightweight\\Ranker};

% Text Node - pAction Ranker (centered)
\draw (913,287) node [anchor=center][inner sep=0.75pt] [align=center] {Final\\Re-Ranker};

% Text Node - Top-K Results (centered)
\draw (1079.5,287) node [anchor=center][inner sep=0.75pt] [align=center] {Top-K Results};

% Text Node - Candidate Generation label
\draw (468,138) node [anchor=center][inner sep=0.75pt] [align=center] {Candidate Generation};

% Text Node - Final Ranking label
\draw (911,202) node [anchor=center][inner sep=0.75pt] [align=center] {Final Ranking};

% % Text Node - GNN features label
% \draw (581,447) node [anchor=center][inner sep=0.75pt] [align=center] {GNN features};

\end{tikzpicture}%
}%
\Description{A pipeline diagram showing the multi-source candidate generation architecture. A reference property feeds into parallel IBKNN and GNN candidate generators, which pass through Business Rules, Merger, and Lightweight Ranker stages within a Candidate Generation box. Output flows to a Final Re-Ranker in the Final Ranking stage, producing Top-K Results.}
\caption{Multi-source candidate generation architecture. CG sources (IBKNN, GNN) generate candidates independently. Business rules (including geo-filtering) are applied, candidates are merged, and lightweight re-ranking produces the final candidate set for the Final Re-Ranker.}
\label{fig:architecture}
\end{figure*}

Figure~\ref{fig:architecture} illustrates our end-to-end candidate generation and ranking pipeline. Given a reference property $P_{ref}$, multiple CG sources generate candidates in parallel. IBKNN retrieves properties with high audience overlap, while the GNN produces candidates based on learned embeddings from a heterogeneous property-user graph. Each source's candidates pass through business rules that enforce hard constraints (geographic proximity, availability, capacity compatibility) before merging. The merger combines candidates from all sources, applying deduplication and score normalization. A lightweight ranker then re-orders the merged candidates based on source confidence and feature-matching signals, producing the final candidate set.

This candidate set is passed to the final re-ranker, a downstream ranking model that scores candidates for final presentation.

\subsection{Individual Candidate Generation Sources}
\textbf{Co-occurrence Methods (IBKNN).} Item-based KNN identifies alternatives based on audience overlap--properties viewed by similar user sets are likely substitutes. For properties $A$ and $B$, Jaccard similarity on user overlap is computed as:
\begin{equation}
\text{sim}_{audience}(A, B) = \frac{|Users_A \cap Users_B|}{|Users_A \cup Users_B|}
\end{equation}
This normalizes for popularity, unlike raw co-occurrence counts. We extend IBKNN with an amenity boost that enhances scores based on amenity similarity while preserving audience overlap as the primary signal. The boost incorporates a confidence adjustment to account for the number of amenities available for comparison---a 100\% match between properties with few amenities is less meaningful than an 80\% match between properties with many amenities. The confidence score is computed as:
\begin{equation}
\text{confidence} = 1 - e^{-\lambda \times \min(|A_1|, |A_2|)}
\end{equation}
where $|A_1|$ and $|A_2|$ are the amenity set sizes and $\lambda$ is a decay rate controlling how quickly confidence increases with amenity count (we use $\lambda = 0.2$). The final score applies the confidence-weighted amenity similarity as a multiplicative boost:
\begin{equation}
\text{score}_{final} = \text{sim}_{audience} \times (1 + \alpha \times \text{sim}_{amenity} \times \text{confidence})
\end{equation}
where $\alpha$ controls the maximum boost (we use $\alpha = 0.2$, allowing up to 20\% score increase). This formulation ensures amenities can only help, never hurt, a property's ranking, while appropriately discounting amenity matches when few amenities are available for comparison.

\textbf{GNN-Based Embeddings.} As described in Section~\ref{sec:background}, our GNN approach models recommendation as link prediction on a heterogeneous graph with node types for reference properties (LHS), candidate properties (RHS), users, interactions, and ground truth labels. The key elements are: (1) \textit{asymmetric embedding learning}--separate LHS and RHS representations capture directional relationships; (2) \textit{supervised training}--the model predicts validated alternatives over a 7-day forward window; and (3) \textit{rich feature incorporation}--structural, location, policy, quality, and amenity features are aggregated through message passing. Neighborhood sampling controls the trade-off between computational cost and information richness, with configurable hop depths and neighbor counts per edge type.

\subsection{Business Rules and Post-Processing}
Before merging candidates from multiple sources, we apply business rules to ensure candidate quality.

\textbf{Hard Constraints.} Candidates failing these criteria are removed: (1) availability--property must be active with sufficient open dates; (2) geographic--within maximum distance from destination (e.g., 40km); (3) quality--minimum guest rating and review count; (4) capacity--guest capacity within acceptable range of reference property.

\textbf{Score Adjustments.} We apply multiplicative boosts for candidates appearing in multiple sources, matching reference-property policies, sharing a neighborhood, or being premier-partner or high-conversion listings.

\subsection{Candidate Merging and Deduplication}
After filtering, we combine candidates from multiple sources via rank-normalized score fusion. Within each source, scores are converted to inverted percentile ranks ($1 - \text{percent\_rank}$) per reference property, placing every candidate on a common $[0, 1]$ scale regardless of the source's native scoring semantics. Candidates from both sources are then joined; for candidates appearing in both, we retain both rank-normalized scores and take their maximum as the final fused score, attributing the candidate to whichever source ranked it higher. We considered alternatives including sequential priority merge (where one source's candidates are preserved and the other only fills remaining slots) and min-max score normalization, but found that rank-normalized max fusion gave the most stable Recall@$K$ curves across the evaluated range without requiring weight tuning or assumptions about cross-source score comparability. The final candidate set (typically $K=300$--$500$) is passed to the lightweight re-ranker before downstream ranking.

\subsection{Lightweight Re-Ranking}
After merging, we apply a lightweight LightGBM ranker to re-order candidates within each reference property's candidate set. This stage bridges the gap between recall-optimized candidate generation and the production ranking model.

\textbf{Training Data Construction.} We construct training examples from historical user interactions. Positive examples are $(P_{\text{ref}}, P_{\text{alt}})$ pairs where a user viewing $P_{\text{ref}}$ subsequently clicked or booked $P_{\text{alt}}$. We assign graded relevance labels: bookings receive label 2, clicks receive label 1. We maintain a 5:1 negative-to-positive ratio to provide sufficient contrast for learning.

\textbf{Negative Sampling Strategy.} Since random negatives produce easy examples, we use four complementary sampling strategies:
\begin{itemize}[leftmargin=*, itemsep=2pt, topsep=2pt]
    \item \textit{Hard negatives} (40\%)--high-scoring candidates (score $\geq 0.7$) with no clicks, teaching the model to distinguish near-misses from true positives.
    \item \textit{Constraint violators} (30\%)--candidates with policy mismatches (e.g., pet policy), large distances, low premier host scores, or poor room configuration matches, reinforcing business rule preferences.
    \item \textit{Medium-quality negatives} (20\%)--candidates with moderate scores ($0.4 \leq \text{score} < 0.7$), providing calibration across the score distribution.
    \item \textit{Random negatives} (10\%)--uniformly sampled candidates for baseline calibration and edge case coverage.
\end{itemize}
This stratified approach ensures the model learns from challenging examples while maintaining calibration across the full candidate distribution.

\textbf{Features.} The ranker uses three feature categories. (1)~\textit{Per-source retrieval scores}: the individual similarity scores from each channel---IBKNN audience-overlap similarity and the GNN embedding (dot-product) similarity---together with their per-reference rank-normalized transforms. These per-source signals are the ranker's primary features. (2)~\textit{Property features}: attributes of both the reference and candidate properties, including ratings, review counts, pricing, and recent booking activity. (3)~\textit{Pairwise features}: relationships between the property pair such as price ratio, rating difference, distance, and policy alignment. We deliberately exclude the single \emph{fused} candidate-generation score--the aggregate cross-source score produced by the merge step of Section~\ref{sec:methodology}--from the feature set, so that the ranker learns from the distinct per-source and property-level signals rather than reproducing the aggregate candidate-generation ordering. The per-source retrieval scores are retained.

\textbf{Model Configuration.} We train a LambdaRank objective optimizing NDCG@100 to align with recall-focused evaluation. Label gains are set to $[0, 3, 10]$ for labels $[0, 1, 2]$, weighting bookings approximately $3\times$ more than clicks. We use gradient boosted trees with 127 leaves, maximum depth 8, and learning rate 0.05, with early stopping on a held-out 20\% validation set split by query groups.

\textbf{Inference.} At serving time, the model scores all candidates per reference property; predictions are used directly, without blending in the fused candidate-generation score. The top~300 candidates are passed to the downstream re-ranker.

\subsection{Final Re-Ranker}
The final re-ranker serves as the final component in our two-stage recommendation system. It is a multi-task neural network trained with a listwise ranking loss. This ranker integrates three primary components: (1) an end-to-end property representation learning network, (2) a user representation module based on historical clicks and bookings, and (3) a search context mapping network that generates a representation for the search context. All of these representations are then concatenated, fed into a shared bottom, and finally task specific MLPs. 

\textbf{Training Data Construction.}
We construct training data from impression logs collected across carousel placements on our platform's websites. Relevance labels are derived from user interactions with displayed properties within a specified forward-looking window (e.g., 7 days). At a high level, we capture three levels of engagement: property interactions (clicks), booking funnel progression milestones (e.g., availability checks, payment page visits), completed bookings, and cancellations. The training lists combine properties from our candidate generation system with impressed properties from the carousels. 

\textbf{Features.}
The features utilized can be grouped into the following main categories: (1) previous user interactions-the properties a user has previously viewed or booked; (2) search context-information pertaining to where a user is going, how long their stay will be, etc; (3) property features-pricing, location, content information, and aggregated interaction statistics among other hand crafted features. Categorical features are encoded as learned embeddings, while numerical features undergo normalization.

\textbf{Inference.}
During production inference, the model processes candidates from the upstream generation system and produces a listwise ranking to generate final traveler recommendations. Real-time business logic filters are applied to exclude inappropriate recommendations, such as inactive or unavailable properties.

%================================================================
\section{Experimental Setup}\label{sec:experiments}
This section describes the dataset, evaluation metrics, and implementation details for our experiments.

\subsection{Dataset}
We conduct experiments on a large-scale vacation rental platform with the following characteristics:

\textbf{Property Catalog.} The dataset contains over 2 million active vacation rental properties spanning diverse property types (apartments, houses, villas, cabins) across global destinations. Properties vary significantly in size (1-10+ bedrooms), amenities (62 distinct amenity types), and price points.

\textbf{User Interactions.} We collect user interaction data including property views, clicks, and bookings over a 90-day period. Interactions are sessionized based on user device identifiers, with each session representing a coherent browsing episode. The interaction graph contains tens of millions of user-property engagement events.

\textbf{Ground Truth Labels.} We use validated alternative property pairs as ground truth for evaluation. These labels are derived from historical user behavior where users who viewed a reference property subsequently booked an alternative property, filtered for geographic and feature compatibility. The labels table contains property pairs where the alternative was a successful conversion, providing supervised signal for training and evaluation.

\textbf{Train/Test Split.} We use a temporal split: training data spans March 1 to May 31, 2025 (3 months), and evaluation is performed on ground truth pairs from June 2025 (1 month). This separation is enforced end-to-end to prevent leakage across the train/test boundary. The June 2025 evaluation pairs are excluded from GNN graph construction, from message passing and neighborhood sampling during training, and from the embeddings that populate the FAISS retrieval index; the GNN is trained solely on interactions and label edges from the March-May window. Property node features are computed from a point-in-time snapshot of the property catalog as of May 31, 2025--including snapshot-aligned review summaries, with trailing-window review counts excluded--so no post-cutoff information is available to the model. The June pairs are used only as held-out targets when scoring retrieval, so recall is measured against alternatives the model never observed during training.

\subsection{Evaluation Metrics}
We evaluate candidate generation quality using the following metrics:

\textbf{Recall@K.} The primary metric measuring what fraction of ground truth alternatives appear in the top-$K$ candidates. We report Recall@$K$ for $K \in \{10, 50, 100, 200, 300, 500\}$ to understand performance across different candidate set sizes.

\textbf{NDCG@K.} As a supporting metric, we report Normalized Discounted Cumulative Gain, which accounts for both the presence and the ranked position of relevant alternatives within the top-$K$ pool. At the candidate-generation stage recall (pool coverage) is the primary objective, since final ordering is established by the downstream ranker; we report NDCG@$K$ alongside recall to characterize within-pool placement.

We additionally measure downstream ranking quality, reporting Booking NDCG@5 of the Final Re-Ranker trained on each candidate pool.

\subsection{Baseline Methods}
We compare against the following candidate generation approaches:

\textbf{Hotel2Vec.} The embedding baseline using Word2Vec-style training on user session sequences. Properties are embedded in a 64-dimensional space, with candidates retrieved via cosine similarity search on a FAISS index.

\textbf{IBKNN.} Item-based collaborative filtering using Jaccard similarity on user audience overlap. For each reference property, we retrieve properties with the highest audience intersection, normalized by union size.

\textbf{Hybrid Baseline.} The existing system combining IBKNN, Hotel2Vec, and distance-based filtering with a LightGBM ranker. This represents our strongest baseline.

\subsection{GNN Implementation Details}
We implement GNN-based candidate generation using Kumo AI's graph learning platform~\cite{fey2024relational} with the following configuration:

\textbf{Graph Construction.} The heterogeneous graph contains five node types: reference properties (hotels\_lhs), candidate properties (hotels\_rhs), users, interactions, and ground truth labels. Edge types connect users to interactions, interactions to properties, and labels to both reference and alternative properties. The separation of properties into LHS and RHS nodes enables asymmetric relationship learning. Consistent with our temporal split (Section~\ref{sec:experiments}), only interactions and label edges from the March--May 2025 window enter the graph; June 2025 pairs are held out for evaluation.

\textbf{Property Features.} Each property node includes: structural features (bedroom\_num, bathroom\_num, sleep\_num, structure\_type), location features (latitude, longitude, market\_id, city\_name), policy features (pets\_allowed, children\_allowed, free\_cancellation, cancellation\_policy), quality features (total\_reviews, average\_rating, premier\_partner status), and 62 one-hot encoded amenity indicators.

\textbf{Training Configuration.} The GNN is trained with a 7-day forward prediction window: given historical interactions, the model learns to predict which properties will serve as validated alternatives in the subsequent week. We train for up to 8 epochs with early stopping (patience=3), using cross-entropy loss with base learning rate 0.01 (reduced to 0.005 in later experiments), weight decay $5 \times 10^{-7}$, and batch size 512. We use cosine learning rate scheduling with warmup (10\% of steps). The GNN architecture uses 128-dimensional hidden channels with multi-aggregation (sum, mean, min, max, std), GELU activation, and layer normalization.

\textbf{Neighborhood Sampling.} For each reference property, we sample neighbors at multiple hops:
\begin{itemize}
    \item Hop 1: 24 label neighbors (supervised signal) + 64 interaction neighbors (behavioral signal)
    \item Hop 2: 1 neighbor per first-hop node for transitive relationship discovery
\end{itemize}
This configuration balances supervised signals from validated alternatives with unsupervised behavioral patterns from user interactions.

\textbf{Embedding Extraction.} After training, we extract dual embeddings per property: LHS embeddings (query representation) and RHS embeddings (candidate representation). We experiment with output dimensions of 32 and 64. Embeddings are used without normalization to preserve dot product semantics.

\textbf{Index Building.} RHS embeddings are indexed using FAISS for efficient similarity search. For each reference property, we query with its LHS embedding and retrieve the top-1000 candidates ranked by dot product similarity. This enables sub-second candidate generation across millions of properties.

\subsection{Model Variants}
We evaluate several GNN configurations varying neighborhood sampling (24 vs.\ 48 label neighbors), output dimension (32 vs.\ 64), and feature set (structural only vs.\ enhanced with amenity and review features). The best configuration uses 48 label neighbors, 64-dimensional embeddings, and the enhanced feature set; all reported GNN results use this model. All experiments use the same protocol: generating top-1000 candidates per reference property and computing metrics against held-out ground truth alternatives.

\subsection{Reproducibility}
We summarize the settings needed to reproduce our results. The GNN is a heterogeneous link-prediction model over five node types (reference and candidate properties, users, interactions, and label edges), trained on the March--May 2025 window with a 7-day forward-prediction target; the architecture uses 128-dimensional hidden channels with multi-aggregation (sum, mean, min, max, std), GELU activations, and layer normalization, and we extract 64-dimensional dual (LHS/RHS) embeddings indexed in FAISS (Section~\ref{sec:experiments}). The lightweight LightGBM ranker is trained with the LambdaRank objective and multi-strategy negative sampling--hard (40\%), constraint-violating (30\%), medium-quality (20\%), and random (10\%)--at a 5:1 negative-to-positive ratio (Section~\ref{sec:methodology}). Its features are the per-source retrieval scores (IBKNN and GNN similarities and their per-reference rank-normalized transforms) together with property and pairwise features; the single fused candidate-generation score is excluded from the feature set. The underlying interaction and catalog data are proprietary, and the GNN is trained on a commercial graph-learning platform, so we release configuration and methodological detail rather than data or model checkpoints.

%================================================================
\section{Results}\label{sec:results}
This section presents our experimental results, comparing GNN-based candidate generation against baselines and examining how candidate-pool quality affects downstream ranking.

\subsection{GNN vs. Embedding Baseline}
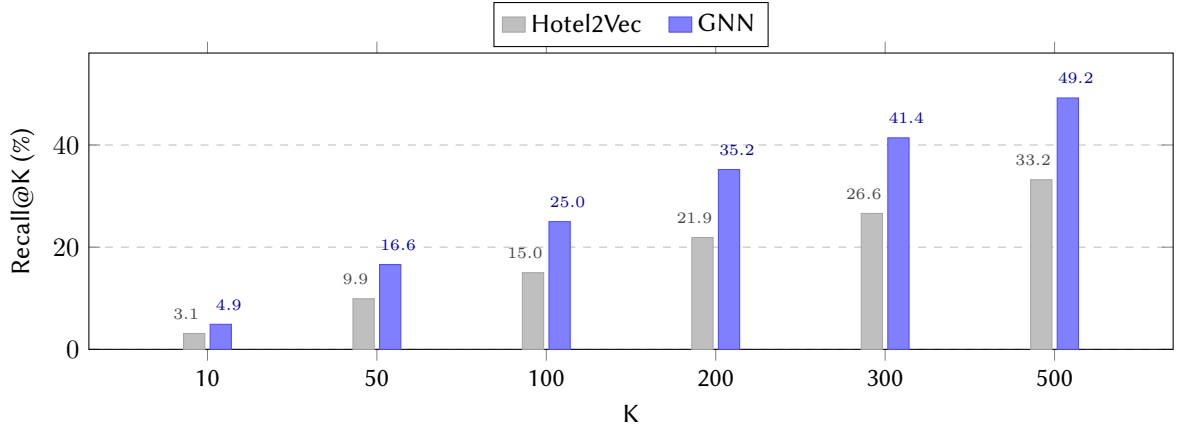
\begin{figure}[t]
\centering
\begin{tikzpicture}
\begin{axis}[
    width=\columnwidth,
    height=5.5cm,
    ybar=2pt,
    bar width=8pt,
    xlabel={K},
    ylabel={Recall@K (\%)},
    ymin=0,
    ymax=58,
    xtick=data,
    xticklabels={10, 50, 100, 200, 300, 500},
    legend style={
        at={(0.5,1.02)},
        anchor=south,
        font=\small,
        legend columns=2,
        /tikz/every even column/.append style={column sep=8pt}
    },
    legend image code/.code={
        \draw[#1] (0cm,-0.1cm) rectangle (0.3cm,0.1cm);
    },
    ymajorgrids=true,
    grid style=dashed,
    enlarge x limits=0.14,
    clip=false,
    nodes near coords,
    nodes near coords style={
        font=\tiny,
        anchor=south,
        yshift=1pt,
        /pgf/number format/fixed,
        /pgf/number format/fixed zerofill,
        /pgf/number format/precision=1
    },
]
% Hotel2Vec data (labels nudged left)
\addplot[fill=gray!50, draw=gray!70,
    nodes near coords style={
        font=\tiny, anchor=south, yshift=1pt, xshift=-3pt, text=gray!55!black,
        /pgf/number format/fixed, /pgf/number format/fixed zerofill,
        /pgf/number format/precision=1
    }] coordinates {
    (1, 3.1)
    (2, 9.9)
    (3, 15.0)
    (4, 21.9)
    (5, 26.6)
    (6, 33.2)
};
% GNN data (labels nudged right)
\addplot[fill=blue!50, draw=blue!70,
    nodes near coords style={
        font=\tiny, anchor=south, yshift=1pt, xshift=3pt, text=blue!65!black,
        /pgf/number format/fixed, /pgf/number format/fixed zerofill,
        /pgf/number format/precision=1
    }] coordinates {
    (1, 4.9)
    (2, 16.6)
    (3, 25.0)
    (4, 35.2)
    (5, 41.4)
    (6, 49.2)
};
\legend{Hotel2Vec, GNN}
\end{axis}
\end{tikzpicture}
\Description{A grouped bar chart comparing Recall@K between Hotel2Vec (gray bars) and the GNN (blue bars) at K values of 10, 50, 100, 200, 300, and 500. The GNN consistently outperforms Hotel2Vec, with values ranging from 4.9\% to 49.2\% compared to Hotel2Vec's 3.1\% to 33.2\%.}
\caption{Recall@K comparison between Hotel2Vec baseline and GNN. GNN achieves 48--68\% relative improvement across all K values, with largest gains at mid-range (K=50, 100).}
\label{fig:recall_comparison}
\end{figure}

Fig.~\ref{fig:recall_comparison} compares the best-performing GNN model against the Hotel2Vec baseline across recall metrics. The GNN approach achieves 48--68\% relative improvement across all recall thresholds, with the largest gains at mid-range K values (Recall@50, Recall@100). This suggests that GNN embeddings are particularly effective at discovering relevant alternatives beyond the most obvious matches.

\subsection{GNN Ablation Studies}
We conduct ablation studies to understand the contribution of different model components.

\textbf{Embedding Dimensions.} Comparing 32- and 64-dimensional output embeddings (holding sampling and features fixed), we observe a small Recall@300 improvement (40.9\% to 41.4\%, +0.5pp). The gain is modest but consistent across $K$, and we adopt 64-dim for all reported results.

\textbf{Neighborhood Sampling.} Increasing label neighbors from 24 to 48 while holding 64 interaction neighbors fixed improves Recall@300 from 39.8\% to 40.5\% (+0.7pp), indicating value in stronger supervised signal. The two-hop sampling strategy is essential for capturing transitive relationships.

\textbf{Feature Importance.} Adding amenity features (62 one-hot indicators) and review features to the base structural features improves Recall@300 from 38.2\% to 39.8\% (+1.6pp), and we adopt the enhanced feature set for all reported results. (This comparison varies output dimension as well; the feature change dominates, as the enhanced model exceeds the basic model despite lower-dimensional embeddings.)

\subsection{Hybrid Architecture Performance}
% Table~\ref{tab:hybrid_cg} compares our proposed hybrid architecture (IBKNN + GNN) against the baseline system (IBKNN + Hotel2Vec).
Table~\ref{tab:hybrid_cg} compares our proposed hybrid architecture (IBKNN + GNN) against the baseline system (IBKNN + Hotel2Vec). Both configurations share the IBKNN source, so this is a like-for-like comparison that isolates the effect of the second source: replacing Hotel2Vec with the GNN in an otherwise identical hybrid.
\begin{table}[t]
\caption{Hybrid CG (after lightweight LightGBM re-ranking): proposed vs.\ baseline system.}
\label{tab:hybrid_cg}
\small
\begin{tabular}{lcccc}
\toprule
\textbf{Config} & \textbf{R@10} & \textbf{R@100} & \textbf{R@200} & \textbf{R@300} \\
\midrule
Baseline & 7.8\% & 29.0\% & 37.1\% & 39.9\% \\
Proposed & 7.9\% & 30.6\% & 40.2\% & 45.8\% \\
\addlinespace
\textit{Rel. Impr.} & +1.3\% & +5.5\% & +8.4\% & +14.8\% \\
\bottomrule
\end{tabular}
\end{table}

The hybrid architecture shows modest improvements at early recall (+1.3\% at Recall@10) but substantial gains at late recall (+14.8\% at Recall@300). This pattern confirms our hypothesis that IBKNN captures obvious alternatives effectively while GNN contributes diverse, non-obvious candidates that expand coverage at higher K values. We note that the GNN is not intended to outperform IBKNN as a standalone source -- indeed, in isolation it trails IBKNN at low $K$ (Table~\ref{tab:complementarity}) -- the contribution lies in the union, where the GNN recovers relevant alternatives that IBKNN alone misses.
% This pattern confirms our hypothesis that IBKNN captures obvious alternatives effectively while GNN contributes diverse, non-obvious candidates that expand coverage at higher K values.

Table~\ref{tab:hybrid_ndcg} reports NDCG@$K$ for the same two configurations and evaluation runs. Recall is the primary candidate-generation objective, since it measures whether relevant alternatives enter the pool passed to the downstream ranker; NDCG@$K$ is reported alongside it to characterize their placement within that pool. The NDCG improvement is positive at every cutoff but smaller than the Recall improvement (+6.6\% vs.\ +14.8\% at $K{=}300$) and concentrated at higher $K$, with early-$K$ NDCG essentially unchanged. This is the expected signature of the GNN's contribution: it enlarges coverage by adding relevant alternatives in the mid-to-late ranks rather than reordering the top of the list, and final positional quality at the top is established by the downstream ranker rather than at the candidate-generation stage. The two metrics are therefore consistent -- the hybrid recovers substantially more relevant alternatives (Recall) while leaving top-of-pool ordering to the ranking stage (NDCG), exactly as intended in a two-stage design.

\begin{table}[t]
\caption{Candidate-generation NDCG@$K$ for the same configurations and runs as
Table~\ref{tab:hybrid_cg}. Recall (Table~\ref{tab:hybrid_cg}) is the primary
candidate-generation metric; NDCG@$K$ is reported alongside it to characterize
within-pool placement.}
\label{tab:hybrid_ndcg}
\small
\begin{tabular}{lcccc}
\toprule
\textbf{Config} & \textbf{N@10} & \textbf{N@100} & \textbf{N@200} & \textbf{N@300} \\
\midrule
Baseline & 0.1067 & 0.2214 & 0.2623 & 0.2764 \\
Proposed & 0.1071 & 0.2219 & 0.2680 & 0.2947 \\
\addlinespace
\textit{Rel. Impr.} & +0.4\% & +0.2\% & +2.1\% & +6.6\% \\
\bottomrule
\end{tabular}
\end{table}

\subsection{Source Complementarity}
To substantiate the division-of-labor claim between the two retrieval sources directly, Table~\ref{tab:complementarity} reports Recall@$K$ for each source in isolation (IBKNN and the GNN) alongside the raw hybrid pool on the June 2025 test set, at the retrieval stage before lightweight re-ranking. IBKNN leads at early recall (28.5\% vs.\ 25.0\% at $K{=}100$), consistent with its strength at surfacing obvious, high-audience-overlap alternatives. The GNN closes the gap and overtakes IBKNN at higher $K$ (49.2\% vs.\ 48.5\% at $K{=}500$), consistent with its role in retrieving diverse, less-obvious alternatives. The hybrid pool matches or exceeds either single source at every cutoff, with a clear margin from $K{=}50$ onward (e.g.\ 44.3\% vs.\ 42.4\% and 41.4\% at $K{=}300$); at $K{=}10$ it equals IBKNN, consistent with IBKNN already supplying the obvious top alternatives while the GNN's contribution emerges at higher $K$. That the union improves over the stronger single source across the range provides direct evidence that the two sources recover partially non-overlapping relevant alternatives rather than redundant ones.

\begin{table}[t]
\caption{Per-source candidate-generation Recall@$K$ (retrieval stage, before lightweight re-ranking) on the June 2025 test set. IBKNN leads at early $K$; the GNN overtakes at high $K$; the fused hybrid matches or exceeds both, with a clear margin from $K{=}50$ onward.}
\label{tab:complementarity}
\small
\begin{tabular}{lcccccc}
\toprule
\textbf{Source} & \textbf{R@10} & \textbf{R@50} & \textbf{R@100} & \textbf{R@200} & \textbf{R@300} & \textbf{R@500} \\
\midrule
IBKNN  & 7.6\%  & 20.6\% & 28.5\% & 37.2\% & 42.4\% & 48.5\% \\
GNN    & 4.9\%  & 16.6\% & 25.0\% & 35.2\% & 41.4\% & 49.2\% \\
Hybrid (fused) & 7.6\%  & 21.2\% & 29.5\% & 38.8\% & 44.3\% & 51.0\% \\
\bottomrule
\end{tabular}
\end{table}

\subsection{Downstream Effect of the Candidate Pool}
We additionally measure how the choice of candidate pool affects downstream ranking quality, holding the ranker architecture fixed. We report our primary offline ranking
metric, Booking NDCG@5, where the relevance label is a binary indicator of whether the user booked the property within the evaluation window, for the test set (Test N@5) and the validation set (Val N@5).

\begin{table}[t]
\caption{Downstream ranking quality by candidate pool. Both
rows use the same Final Re-Ranker with an identical feature
set and architecture, trained and evaluated on its own
candidate pool.}
\label{tab:e2e}
\small
\begin{tabular}{lcc}
\toprule
\textbf{Candidate Pool} & \textbf{Test N@5} & \textbf{Val N@5} \\
\midrule
Baseline (IBKNN + Hotel2Vec) & 0.345 & 0.361 \\
Hybrid, proposed (IBKNN + GNN) & \textbf{0.365} & \textbf{0.369} \\
\bottomrule
\end{tabular}
\end{table}

As shown in Table~\ref{tab:e2e}, the hybrid candidate pool yields higher downstream Booking NDCG@5 than the baseline pool. We note that each ranker is trained on its own pool, so this comparison should be read as the combined effect of pool composition and ranker retraining; isolating the two is left to future work.

%================================================================
\section{Discussion}\label{sec:discussion}
Our results offer several insights for practitioners building two-stage
recommendation systems over heterogeneous, long-tail inventory.

\textbf{Why GNN candidates help most at higher $K$.}
The hybrid architecture shows only a marginal gain at Recall@10
(+1.3\%) but a substantial one at Recall@300 (+14.8\%). We attribute
this pattern to the division of labor between the two sources. IBKNN,
driven by audience overlap, reliably surfaces the most obvious
alternatives -- properties co-viewed by large, overlapping user
sets -- and these dominate the early ranks. The GNN contributes
candidates that audience overlap alone does not reach: properties
connected through multi-hop behavioral paths or through shared
attributes rather than shared audiences. These are not the first
properties a popularity-driven method would return, so their effect
is concentrated in the mid-to-late ranks where the candidate pool
would otherwise thin out. For a candidate-generation stage whose job
is to maximize coverage before ranking, this late-recall expansion is
the operative contribution.

\textbf{Cold-start and the role of explicit features.}
A recurring theme is the GNN's ability to represent properties with
sparse interaction history. IBKNN and Hotel2Vec both degrade for
cold-start listings: the former has little audience overlap to
measure, the latter little co-view signal to embed. Because the GNN
aggregates explicit property features -- structure, location, policy,
quality, and amenities -- through message passing, it can place a
new property in a sensible region of the embedding space even before
that property accumulates interactions. Our ablations support this:
adding amenity and review features yields a +1.6pp absolute
Recall@300 improvement, indicating that explicit features and
behavioral signal are complementary rather than redundant.

\textbf{The recall-conversion gap.}
A central practical finding is that gains at the candidate-generation
stage do not transfer to the ranking stage in a simple or directly
attributable way. The hybrid candidate pool yields a higher
downstream Booking NDCG@5 than the baseline pool (0.365 vs.\ 0.345
on test), but this measurement confounds two effects: the change in
pool composition and the retraining of the ranker on that new pool.
A ranker trained on one candidate distribution and evaluated on
another is not a clean comparison, and we deliberately do not claim
one. The broader lesson is that introducing a new CG source is not a
drop-in change: the downstream ranker is tuned, implicitly or
explicitly, to the distribution of candidates it was trained on, and
realizing the full value of a better candidate pool requires
co-adapting the ranking stage. Practitioners should budget for this
coupling rather than expecting retrieval gains to convert
automatically.

\textbf{Limitations.} We state the scope of the study explicitly. \emph{Attribution of the hybrid gain.} The proposed and baseline systems in Table~\ref{tab:hybrid_cg} differ in more than one respect: the proposed pool replaces the Hotel2Vec source with the GNN \emph{and} uses rank-normalized max fusion in place of the baseline's sequential source-priority merge. The +14.8\% Recall@300 should therefore be read as the combined effect of the source and the fusion change rather than as an isolated measure of the GNN source. \emph{Baseline supervision.} The GNN is trained on booking-derived alternative pairs with rich property features, whereas Hotel2Vec is a co-view embedding with different supervision; the comparison thus reflects both representation and supervision differences and does not isolate the effect of graph message passing relative to a supervised feature-based retriever. 
\emph{Complementarity and cold start.} We report per-source Recall@$K$ (Table~\ref{tab:complementarity}) that directly establishes the IBKNN-early / GNN-late division of labor; however, we do not report source overlap, unique-hit, or oracle-union analyses, nor recall stratified by interaction-degree or listing age. In particular, although cold-start handling motivates our use of the GNN, we do not include a dedicated ablation measuring recall on newly onboarded or low-interaction properties; we therefore treat the cold-start benefit as motivated but not yet directly quantified, and regard such an ablation as a necessary next step rather than deferred future work.
% \emph{Complementarity and cold start.} We report per-source Recall@$K$ (Table~\ref{tab:complementarity}) that directly establishes the IBKNN-early / GNN-late division of labor; however, we do not report source overlap, unique-hit, or oracle-union analyses, nor recall stratified by interaction-degree or listing age, which would further substantiate the cold-start claim in particular. 
\emph{Diversity.} Relatedly, our characterization of GNN candidates as diverse and non-obvious is qualitative; we do not report catalog-coverage, novelty, or long-tail exposure metrics. \emph{Exposure bias.} Our ground truth is derived from observed booking conversions, which inherits the exposure bias of the deployed system: alternatives never surfaced cannot appear as positives, so recall is measured against a partially observed target. \emph{Statistical reporting and downstream attribution.} We evaluate on a single monthly window without confidence intervals, so the smaller differences (e.g., +1.3\% at Recall@10) should be read cautiously; and the downstream Booking NDCG@5 comparison retrains the ranker on each candidate pool, so it reflects the combined effect of pool composition and ranker retraining rather than the candidate pool in isolation. Finally, all results are offline; we report no online (A/B) evaluation, so the recall and ranking gains should be read as offline evidence whose translation to user-facing outcomes remains to be confirmed. The GNN also incurs higher graph-construction and training cost than the shallow baselines, though candidate generation remains sub-second at serving time.
% Finally, the downstream evaluation is offline by design and online behavior may differ, and the GNN incurs higher graph-construction and training cost than the shallow baselines, though candidate generation remains sub-second at serving time.

%================================================================
\section{Conclusion and Future Work}\label{sec:conclusion}
We presented a study of candidate generation for alternative vacation rental recommendations on a large-scale platform with over 2M active properties. Comparing collaborative filtering, shallow embeddings,
and graph neural networks, we found that GNN-based retrieval
substantially outperforms a Hotel2Vec baseline (48--68\% relative
recall improvement), and that a hybrid architecture combining IBKNN
with a GNN source improves Recall@300 by +14.8\% over the strongest
baseline by pairing IBKNN's early-recall strength with the GNN's
late-recall diversity. We further observed that the stronger hybrid
candidate pool carries through to higher downstream ranking quality,
while noting that cleanly attributing this gain is complicated by the
coupling between candidate generation and ranker training -- the recall-conversion gap.

Several directions follow from this work. First, a controlled, one-variable-at-a-time evaluation would attribute the hybrid gain precisely: replacing only the source (Hotel2Vec $\rightarrow$ GNN) with fusion held fixed, and comparing the GNN against a supervision-matched two-tower retriever and graph baselines such as LightGCN and PinSage, would isolate the contribution of graph message passing. Second, a fixed-ranker (cross-pool) downstream evaluation that holds the ranker's training distribution fixed would separate candidate-pool quality from ranker retraining and quantify the recall-conversion gap rather than only observing it. Third, per-source overlap, unique-hit, and oracle-union analyses, together with recall stratified by interaction-degree and listing age, would directly test the complementarity and cold-start claims, while catalog-coverage, novelty, and long-tail metrics would substantiate the diversity claim. Fourth, reporting over multiple temporal windows with bootstrap confidence intervals, and mitigating exposure bias via inverse-propensity weighting or a small randomized-exposure bucket, would strengthen the statistical basis of the offline results. Finally, an online evaluation would test whether the offline recall and ranking improvements translate into user-facing conversion gains, and a learned, parameterized fusion function over per-source rank-normalized scores (rather than the unweighted max used here) may yield further coverage gains.

%% =====================================================================
\section*{Declaration on Generative AI}
In preparing this paper, we used Claude (Anthropic) for editorial assistance on the initial draft, including correcting writing errors, refining prose for clarity and conciseness, and suggesting revisions to improve structure and remove redundancy. No AI tools were used for data analysis, experimentation, or the formulation of conclusions. All AI-assisted edits were reviewed and approved by the authors, who take full responsibility for the final content.

%% =====================================================================
%% Bibliography. Confirmed against the CEUR sample-1col.tex: it uses a bare
%% \bibliography{...} with NO \bibliographystyle (the ceurart class sets the
%% style). Keep your sample-base.bib in the project.
%% =====================================================================
\bibliography{sample-base}

\end{document}